\documentclass[11pt]{article}

\usepackage[preprint]{acl}

\usepackage{times}
\usepackage{latexsym}

\usepackage[T1]{fontenc}

\usepackage[utf8]{inputenc}

\usepackage{microtype}

\usepackage{inconsolata}

\usepackage{graphicx}

\usepackage{amsmath}
\usepackage{amsthm}
\usepackage[table]{xcolor}
\usepackage{adjustbox}
\usepackage{makecell}
\usepackage{booktabs}
\usepackage{multirow}
\usepackage[ruled,vlined,linesnumbered]{algorithm2e}
\usepackage{amssymb}
\usepackage{subcaption}

\title{PPL-Factory: Task-Aware and Budget-Aware Data Selection from Language Modeling to Reasoning}

\author{Hang Zhang \qquad Warren J. Gross \\
    Department of Electrical and Computer Engineering, McGill University \\ Montreal, QC, Canada\\
    \small{
        \textbf{Correspondence:} \href{mailto:hang.zhang3@mail.mcgill.ca}{hang.zhang3@mail.mcgill.ca}}}

\begin{document}
\maketitle
\begin{abstract}
Not all training samples contribute equally to large language model fine-tuning.
Selecting informative training samples can reduce the computational cost while preserving downstream performance.
Many existing data selection methods rely on indirect heuristics, such as data quality, diversity or reasoning trace length.
However, the effectiveness of these fixed criteria is task-dependent and difficult to generalize across diverse downstream tasks.
Perplexity-based data selection provides a simple and model-aware solution to estimate the sample difficulty, but existing approaches typically score the entire training sequence and ignore the difference in learning objectives of language modeling and reasoning tasks.
In this paper, we propose \textbf{PPL-Factory}, a simple and interpretable data selection framework that combines task-aware perplexity-based scores and data budget-aware selection criteria.
Experiments on GSM8K demonstrate that PPL-Factory outperforms other state-of-the-art data selection methods using only $1\%$ of the training set.
With $10\%$ of the data, PPL-Factory exceeds full-data fine-tuning accuracy by 0.9 on GSM8K and 4.8 on MATH.
Overall, our results demonstrate that task-aware and budget-aware perplexity-based selection provides an effective and applicable approach for efficient fine-tuning.
\end{abstract}

\section{Introduction}
Fine-tuning large language models (LLMs) has become a standard approach for adapting pretrained models on diverse downstream tasks.
However, full dataset fine-tuning remains computationally expensive, especially when the parameter size for the model or the available training corpus is large~\cite{albalak2024a}.
Considering that not all training examples give equal learnable signals to the model improvement, data selection and data pruning approaches~\cite{chen2024alpagasus, li-etal-2024-superfiltering} are proposed to discard redundant, overly hard, noisy or mislabeled samples for LLMs.
Identifying a subset of training data that preserves the informative knowledge is therefore an active research topic.

Perplexity offers a simple and model-aware signal for estimating the difficulty of training examples.
Recent studies have shown that both high and medium perplexity improve performance and training efficiency~\cite{ankner2025perplexed}.
However, prior work has shown that many one-shot coreset selection methods suffer from a dramatic accuracy drop under extremely low data-budget settings and can even perform worse than random sampling~\cite{zheng2023coveragecentric, xia-etal-2025-rethinking}, due to insufficient data diversity and informative coverage.
Embedding-based data selection approaches~\cite{rubin-etal-2022-learning, ni-etal-2022-large} utilize a proxy model to assign a score to each data point in the training set based on its similarity with the validation set.
However, semantic similarity extracted from a separate frozen model has a weak connection to the learning behavior of the target model.
Gradient similarity-based approaches~\cite{10.5555/3692070.3694291, NEURIPS2025_d2d4f685} evaluate the utility of each data point and compute its training gradient.
Influence functions~\cite{10.5555/3305381.3305576, dai-etal-2025-improving} estimate how much a training example affects a model prediction without fully retraining the model.
Despite their promising performance, these sophisticated scores often introduce additional computational overhead, as they may require gradient construction, gradient-feature storage, Hessian-vector products, or influence estimation~\cite{Hammoudeh_2024, pmlr-v139-killamsetty21a}.
The utility functions and computational complexities for the previously mentioned approaches are compared in Table~\ref{tab:selection_cost}.

\begin{table*}[t]
  \centering
  \begin{tabular}{lcc}
    \hline
    \textbf{Method} & \textbf{Utility Function} & \textbf{Per-example Computational Cost} \\
    \hline
    Random & $u(x) \sim \mathcal{U}(0,1)$ & $\mathcal{O}(1)$ \\
    Perplexity-Based & $u(x)=\mathrm{PPL}_{\theta}(x)$ & $C_{\mathrm{forward}}(x)$ \\
    Embedding-Based & $u(x)=\mathrm{sim}\!\left(h_{\theta}(x), h_{\theta}(\mathcal{V})\right)$ & $C_{\mathrm{forward}}(x)+\mathcal{O}(d|\mathcal{V}|)$ \\
    Gradient-Based & $u(x)=\left\|\nabla_{\theta}\ell_{\theta}(x)\right\|$ & $C_{\mathrm{forward}}(x)+C_{\mathrm{backward}}(x)$ \\
    Influence-Based & $u(x)=-\nabla_{\theta}\ell_{\theta}(\mathcal{V})^{\top}H_{\theta}^{-1}\nabla_{\theta}\ell_{\theta}(x)$ & $C_{\mathrm{forward}}(x)+C_{\mathrm{backward}}(x)+C_{\mathrm{IHVP}}$ \\
    \hline
  \end{tabular}
  \caption{Utility functions and per-example computational cost for data selection methods.
  Random selection as a common strong baseline~\cite{ivison2025largescaledataselectioninstruction, zheng2023coveragecentric} has the lowest computational complexity.
  Perplexity scoring computation only requires forward passes without gradient updates.
  $h_{\theta}(x)\in\mathbb{R}^{d}$ denotes the embedding of the training example $x$, $\mathcal{V}$ denotes a small validation set.
  Embedding-based approaches need a one-time preprocessing cost on validation embeddings.
  $\ell_{\theta}(x)$ denotes the token-level negative log-likelihood.
  Both gradient-based and influence-based approaches involve a backward pass, and influence-based utility further requires the inverse Hessian $H_{\theta}^{-1}$.}
  \label{tab:selection_cost}
\end{table*}

To address the above limitations, we propose \textbf{PPL-Factory}, a task-aware and data budget-aware framework for LLM efficient fine-tuning.
Unlike prior perplexity-based methods that mainly rely on deterministic full-sequence difficulty ranking, our method extends the perplexity-based selection from directly aligned language modeling tasks to reasoning tasks and applies a budget-aware selection criterion.
For causal language modeling (CLM) tasks, PPL-Factory scores fixed-length text blocks using token-level negative log-likelihood (NLL).
As for the reasoning tasks, where the informative signal is not always uniformly distributed across the full sequence, PPL-Factory adapts a weighted combination of reasoning NLL score and answer NLL score.
In addition, we study the change of the best selection criterion under diverse data budgets, and propose a budget-aware selection criterion to preserve the model performance, especially under extremely low data budgets.
Our main contributions can be summarized as follows:
\begin{itemize}
    \item We propose \textbf{PPL-Factory}, a joint task-aware and budget-aware framework for efficient LLM fine-tuning.
    PPL-Factory selects informative training subsets based on likelihood within the target model.
    Experimental results show that our approach surpasses the state-of-the-art methods on GSM8K only using $1\%$ of the training set.
    It provides a lightweight alternative to existing selection methods that require external evaluators or expensive quality annotations.

    \item We develop task-aware NLL-based selection scores for language modeling and reasoning tasks within the PPL-Factory framework.
    Instead of applying conventional full-sequence perplexity ranking, the proposed scores compute NLL over task-characterized regions.
    Across four widely used benchmarks, our method consistently outperforms prior perplexity-based selection when using less than $30\%$ of the training set, highlighting the robustness and generalizability under low data budgets.

    \item We systematically evaluate the fine-tuning performance of PPL-Factory across different training data budgets.
    The results show that the proposed budget-aware selection strategy outperforms full-data fine-tuning on reasoning tasks and achieves comparable performance on CLM tasks while using substantially fewer training examples.
\end{itemize}

\section{Preliminaries}
\label{sec:preliminaries}
\paragraph{Negative Log-Likelihood and Perplexity}
NLL measures how unlikely a sequence is under a language model.
Given a token sequence $\mathbf{x}=(x_1, x_2\ldots,x_T)$, the token-normalized NLL is defined as
\begin{equation}
    s_{\mathrm{NLL}}(x_i;\theta) = -\frac{1}{T}\sum_{t=1}^{T}\log p_{\theta}(x_t \mid x_{i,<t}).
\end{equation}

Perplexity is the exponential form of the token-normalized NLL
\begin{equation}
    \mathrm{PPL}(x_i;\theta) = \exp\left(s_{\mathrm{NLL}}(x_i;\theta)\right).
\end{equation}
A lower NLL means that the language model predicts the next token relatively well and confidently, so it reflects an easy sample in the set.
Middle-band NLL samples provide more training updates, and they are still learnable.
High values for NLL mean the samples have surprise learning signals to the model, thus are difficult to learn based on pre-trained parameters.
Since perplexity is directly derived from the training objective of autoregressive language models, it provides a model-aware measure of sample difficulty.

\paragraph{One-shot Data Selection}
Let $\mathcal{D}=\{z_i\}_{i=1}^N$ denote a training dataset with $N$ examples, where each $z_i$ represents a training unit.
One-shot data selection aims to select a training subset $\mathcal{D}_\rho \subseteq \mathcal{D}$ before training, given a selection ratio $\rho \in (0, 1]$, such that the evaluation performance on the target dataset can be preserved.
This optimization objective can be formulated as
\begin{equation}
  \begin{aligned}
    \arg\min_{\mathcal{D}_{\rho}\subseteq \mathcal{D}} \; \mathcal{L}\left(\theta^{*}(\mathcal{D}_{\rho}); \mathcal{D}_{\mathrm{eval}}\right), \\
    \mathrm{s.t.} \quad |\mathcal{D}_{\rho}|=\lfloor \rho N \rfloor,
  \end{aligned}
\end{equation}
where $\mathcal{L}(\cdot)$ denotes the loss function, and 
$\theta^{*}(\mathcal{D}_{\rho})$ denotes the model parameters trained on the selected subset $\mathcal{D}_{\rho}$.

\paragraph{Supervised Fine-tuning.}
Given a certain dataset containing target sequences, the training strategy is to minimize the NLL of the target sequence $y$ given an input $x$, which is,
\begin{equation}
    \mathcal{L}_\text{SFT}(\theta) = -\sum_{t=1}^T \log p_\theta(y_t \mid x,y_{<t}),
\label{eq:NLL_base}
\end{equation}
where $y_{<t} = (y_1, \cdots, y_{t-1})$ denotes the prefix before the $t$-th token in the target response.

\section{Methodology}
\label{sec:method}
\begin{figure*}[t]
  \includegraphics[width=\linewidth]{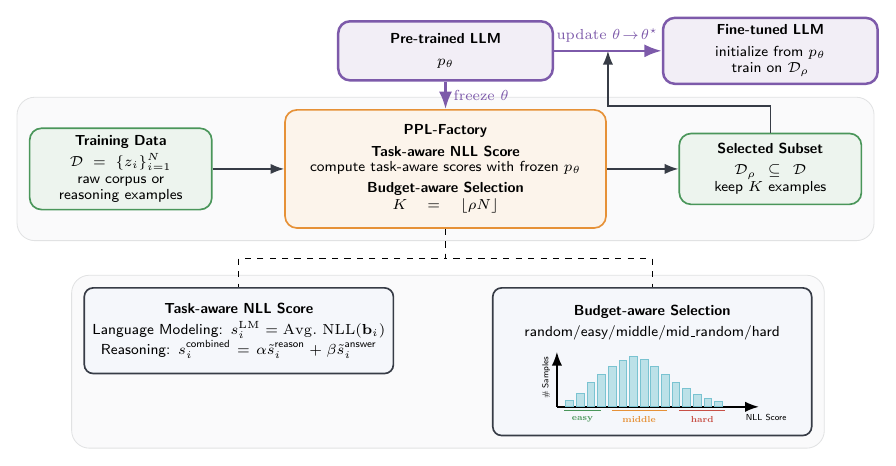}
  \caption{Overview of the proposed PPL-Factory framework.
  The orange-shaded block represents PPL-Factory.
  Purple-shaded blocks represent LLMs, green-shaded blocks represent datasets.
  The top, middle, and bottom rows illustrate the model fine-tuning, data selection flow, and internal components of PPL-Factory, respectively.}
  \label{fig:overview}
\end{figure*}
Figure~\ref{fig:overview} provides an overview of our proposed PPL-Factory.
Given a training corpus, PPL-Factory computes task-aware NLL scores using the frozen pre-trained LLM and applies a budget-aware selection rule to obtain a subset for efficient fine-tuning.
Then the subset is used to fine-tune the pre-trained LLM to improve the downstream performance.

In the following subsections, we first formulate the data selection problem in Section~\ref{sec:method_sub_problem}.
We then introduce the task-aware adaptation to the NLL scores for diverse downstream tasks in Section~\ref{sec:method_sub_task_aware}, from CLM to mathematical reasoning tasks.
Finally, Section~\ref{sec:method_sub_budget_aware} presents our budget-aware selection criteria.

\subsection{Problem Definition}
\label{sec:method_sub_problem}
We consider an offline, one-shot data selection approach for efficient fine-tuning.
Let the original training corpus be denoted as $\mathcal{D}$ and let $\rho \in (0, 1]$ denote the data selection ratio.
The goal is to construct a subset $\mathcal{D}_{\rho}$ which is selected from the full-size dataset $\mathcal{D}$ according to a certain criterion, where $\mathcal{D}_{\rho}$ contains a fraction $\rho$ of the available training samples.
Thus, the selection criterion is to identify informative and learnable signals from the original training set and maximize the training utility of the selected subset.

Let $s$ denote the NLL of a training example, which serves as a model-aware difficulty signal for data selection.
For a given selection ratio $\rho$, we first define the expected utility of retaining an example with score $s$ as
\begin{equation}
    u_\rho(s) = r_\rho(s) - \gamma_\rho c_\rho(s),
    \label{eq:expected_utility}
\end{equation}
where $r_\rho(s)$ is the positive training contribution to fine-tuning and $c_\rho(s)$ denotes the potential negative effects caused by noise, incorrect label, or outlier difficulty.
The coefficient $\gamma_\rho$ controls the penalty assigned to the negative effect.

While Equation~\ref{eq:expected_utility} characterizes the utility of individual examples, an effectively selected subset should keep sufficient coverage of the original distribution.
Let $P(s)$ denote the NLL distribution of the full train set, and let $Q_\rho(s)$ denote the selection score distribution of the selected subset under selection ratio $\rho$.
We formulate data selection as an informative-coverage maximization problem, where the selected subset is encouraged to retain useful training signals while maintaining distributional coverage with respect to the full corpus
\begin{equation}
  \begin{aligned}
    J_\rho(Q_\rho) &= \int u_\rho(s) Q_\rho(s) \, ds \\
    &- \lambda_\rho D_{\mathrm{KL}}(Q_\rho \parallel P), & \lambda_\rho > 0.
  \end{aligned}
  \label{eq:score_informative_coverage}
\end{equation}
The first term measures the expected utility of the selected subset, while the KL-divergence term penalizes excessive deviation from the original score distribution.
The parameter $\lambda_\rho$ controls the trade-off between concentrating on a high-utility score region and increasing coverage of the full-size train set.
Overall, the optimization problem of the data selection for efficient fine-tuning is now transferred to maximizing the informative-coverage score under a selection ratio $\rho$ and $\lambda_\rho > 0$.
For a fixed ratio $\rho$, the maximizer of Equation~\ref{eq:score_informative_coverage} is
\begin{equation}
    Q_\rho^*(s) = \frac{P(s) \exp(u_\rho(s)/\lambda_\rho)}{\int P(t) \exp(u_\rho(t)/\lambda_\rho) \, dt}.
  \label{eq:informative_coverage_maximizer}
\end{equation}
The dependence of $Q_\rho^*(s)$ on $u_\rho$ shows the optimal subset distribution is budget-dependent.
In other words, the beneficial score region may change as the selection ratio $\rho$ varies.
This observation motivates our data budget-aware selection criterion, described in Section~\ref{sec:method_sub_budget_aware}.

\subsection{Task-Aware NLL Scores}
\label{sec:method_sub_task_aware}
Since perplexity directly reflects how likely a language model considers a sequence under its learned distribution, it is directly aligned with the next-token prediction objective.
Motivated by this alignment, we use NLL as a practical scoring form of perplexity.
We then adapt the NLL computation to reasoning task structures, so that the score captures the most relevant learning signal for different downstream tasks.

\paragraph{Causal Language Modeling Tasks}
In CLM fine-tuning, the training corpus is typically tokenized, concatenated, and divided into fixed-length blocks.
Therefore, we pack the corpus into token blocks $\mathbf{b}_i = (b_{i,1},\dots,b_{i,B})$ where $B$ is the block size.
Then we calculate NLL using the average next-token prediction loss over all tokens in a packed block.
Since all packed blocks have the same length, the resulting NLL-based score provides a comparable block-level measure of difficulty.
For each block $\mathbf{b}_i$, the language modeling score is computed as
\begin{equation}
    s_i^\text{LM}=\frac{1}{B-1}\sum_{t=1}^{B-1} -\log p_\theta(b_{i,t+1}\mid b_{i,1:t}).
    \label{eq:nll_modeling}
\end{equation}
Based on the score, we rank the block by difficulty.
A subset $\mathcal{D}_\rho$ can be selected given the selection criterion and the selection ratio.
A detailed description of the proposed automatic selection approach is provided in Algorithm~\ref{alg:nll_blockwise}.

\begin{algorithm}[t]
\caption{Packed-block Data Selector}
\label{alg:nll_blockwise}
\DontPrintSemicolon
\KwIn{Dataset $\mathcal D$, language model $\mathcal M$, block size $B$, selection ratio $\rho$, \texttt{selection} $\in$ \{\texttt{random}, \texttt{easy}, \texttt{hard}, \texttt{middle}, \texttt{mid\_random}\}, lower quantile $q_l$, upper quantile $q_u$}
\KwOut{Selected packed-block dataset $\mathcal D_\rho$}

Tokenize $\mathcal D$, concatenate tokens, and split into blocks $\mathcal B=\{\mathbf b_i\}_{i=1}^{N}$ of length $B$\;
$K\leftarrow \max(1,\lfloor \rho N\rfloor)$\;

\If{$\texttt{selection}=\texttt{random}$}{
    Randomly select $K$ blocks\;
}
\Else{
    $s_i\leftarrow \frac{1}{B-1}\sum_{t=1}^{B-1}-\log p_{\mathcal M}(b_{i,t+1}\mid b_{i,1:t})$, $\forall \mathbf b_i\in\mathcal B$\;

    \uIf{$\texttt{selection}=\texttt{easy}$}{select $K$ blocks with lowest $s_i$\;}
    \uElseIf{$\texttt{selection}=\texttt{hard}$}{select $K$ blocks with highest $s_i$\;}
    \uElseIf{$\texttt{selection}=\texttt{middle}$}{select $K$ blocks closest to $\mathrm{median}(\{s_i\})$\;}
    \ElseIf{$\texttt{selection}=\texttt{mid\_random}$}{
        $\mathcal D'\leftarrow\{\mathbf b_i:Q_{q_l}(\{s_i\})\le s_i\le Q_{q_u}(\{s_i\})\}$\;
        sample $K$ blocks from $\mathcal D'$, with fallback to $\mathcal B\setminus\mathcal D'$\;
    }
}

\Return selected blocks as $\mathcal D_\rho$\;
\end{algorithm}

\paragraph{Mathematical Reasoning Tasks}
Unlike CLM corpora, the learning signal in the reasoning benchmarks is not uniformly distributed over the full sequence.
Reasoning datasets are typically organized as instruction-response pairs, where the response contains the main task-specific supervision for fine-tuning~\cite{huerta-enochian-ko-2024-instruction}.
Both the chain-of-thought reasoning and the final answer are closely related to reasoning performance~\cite{10.5555/3600270.3602070, fang2025what}.

Instead of computing NLL over fixed-length packed blocks, we compute it over the reasoning steps and the final answer.
For each structured example $(q_i, y_i)$, $q_i$ denotes the question and $y_i$ denotes the target response.
The response is further decomposed as $y_i = [y_i^\text{reason}, y_i^\text{answer}]$.
The reasoning and answer NLL scores are computed as
\begin{equation}
 \begin{aligned}
    s_i^\text{reason} &= -\frac{1}{|y_i^\text{reason}|} \sum_{t \in y_i^\text{reason}} \log p_\theta (y_{i,t} \mid q_i, y_{i,<t}), \\
    s_i^\text{answer} &= -\frac{1}{|y_i^\text{answer}|} \sum_{t \in y_i^\text{answer}} \log p_\theta (y_{i,t} \mid q_i, y_{i,<t}).
 \end{aligned}
\end{equation}
We define the overall selection score as a normalized weighted combination of average token-level NLL scores
\begin{equation}
    s_i^{\text{combined}} = \alpha \tilde{s}_i^\text{reason} + \beta \tilde{s}_i^\text{answer}, \\
  \label{eq:nll_reasoning}
\end{equation}
where $\alpha$ and $\beta$ are the controlled weights to the reasoning steps and the final answer.
$\tilde{s}_i$ is the normalized score using z-score normalization.
This design measures the model's uncertainty in generating the solution process and deriving the final answer from the question, thereby producing a more task-aware selection score for mathematical reasoning fine-tuning.

Instead of applying fixed lower and higher quantile thresholds as in Algorithm~\ref{alg:nll_blockwise}, we construct a selection pool whose size is $\texttt{mid\_pool\_ratio}\times$ the target number of selected samples and then randomly sample from this pool.

\subsection{Budget-Aware Selections}
\label{sec:method_sub_budget_aware}
Motivated by Equation~\ref{eq:informative_coverage_maximizer}, we propose a budget-aware data selection strategy based on the scores described in Section~\ref{sec:method_sub_task_aware}.
When the selection ratio is relatively high, we use easier samples to avoid extremely difficult or noisy outliers.
As the budget becomes more limited, we shift the selection toward mid-range samples, which provide more informative learning signals while avoiding both overly simple and overly difficult examples. 
For very small budgets, we further introduce randomness into the middle-band selection to preserve coverage of the training distribution and reduce the risk of selecting a narrow, biased subset. 
This randomization mitigates score noise and improves diversity without introducing additional scoring or evaluator costs.
More details and analysis are discussed in Appendix~\ref{sec:appendix_maximizer}.

\section{Experiments and Results}
\label{sec:exp_results}
In this section, we present a comprehensive validation of our task-aware and budget-aware PPL-Factory.
We first detail our model configuration and evaluation benchmarks in Section~\ref{sec:exp_results_sub_models_benchmarks}.
Then, we list the training details in Section~\ref{sec:exp_results_training}.
In Section~\ref{sec:exp_results_sub_main_results} and Section~\ref{sec:exp_results_sub_ablation}, we compare our results on multiple datasets against baselines and analyze key components through ablation studies.
Finally, we analyze the fine-tuning cost and calculation complexity in Section~\ref{sec:exp_results_sub_complexity}.

\subsection{Benchmarks and Models}
\label{sec:exp_results_sub_models_benchmarks}
We evaluate PPL-Factory on four datasets, including CLM tasks and mathematical reasoning tasks.

In the experiments on CLM tasks, we fine-tune LLaMA3.2-1B, LLaMA3.2-3B, and LLaMA3.1-8B~\cite{grattafiori2024llama3herdmodels} on our selected subset from WikiText-2 and WikiText-103~\cite{merity2016pointer} to evaluate the model performance.
The reported metric is perplexity.

For reasoning tasks, we adopt LLaMA3.2-1B-Instruct, LLaMA3.2-3B-Instruct, and LLaMA3.1-8B-Instruct as backbone models.
We validated our PPL-Factory on two mathematical benchmarks, GSM8K~\cite{cobbe2021trainingverifierssolvemath} and MATH~\cite{hendrycksmath2021}.
Performance is measured by zero-shot exact match accuracy on GSM8K and greedy Pass@1 accuracy on MATH.
Each problem is evaluated using a single deterministic generation.

\subsection{Training Details}
\label{sec:exp_results_training}
All experiments use full parameter fine-tuning and are conducted on a single NVIDIA H100 with 80G memory.
We follow standard practice in LLM fine-tuning~\cite{wolf-etal-2020-transformers}.
For the data selectors, we keep $\texttt{block\_size}=2048$ in the packed-block data selector.
Lower quantile and upper quantile are $q_l=0.1$ and $q_u=0.8$, respectively.
These values are selected to keep more learnable, easier samples and remove more outliers.
In reasoning task settings, we have $\texttt{max\_length} = 2048$ for GSM8K, and $\texttt{max\_length} = 4096$ for MATH since its problems and solutions are much longer and more complex.
The weights in Equation~\ref{eq:nll_reasoning} are $\alpha = 1, \, \beta=0.5$.
This choice assigns primary importance to the reasoning NLL score because it captures the model's intermediate multi-step problem-solving ability.
The answer-level NLL is used as an auxiliary term to count the difficulty of generating the final answer, but with a smaller weight to avoid overemphasizing short answer spans.
$\texttt{mid\_pool\_ratio}=2.0$ when $\rho \leq 0.2$ and $\texttt{mid\_pool\_ratio}=1.5$ when $\rho$ is larger to keep the selection algorithm focusing on the middle band instead of introducing much randomness.

For smaller data budget ($1\%$-$10\%$) experiments, we conduct five independent runs with different random seeds.
For larger data budget ($20\%$-$100\%$) experiments, we perform three independent runs.
To ensure fair comparison, we use consistent optimization seeds across all experiments.
We report the mean value of the evaluation metric in our analysis.
The fine-tuning settings are discussed in Appendix~\ref{sec:appendix_repro_sub_finetuning}.

\subsection{Results}
\label{sec:exp_results_sub_main_results}
Table~\ref{tab:main_results} reports the evaluation performance on the four benchmarks under low data budget settings, with selection ratios ranging from $10\%$ to $30\%$.
We focus on this regime because preserving model performance with limited training data is particularly challenging and is critical to evaluating the effectiveness of data selection methods.
Table~\ref{tab:results_clm} on the left illustrates the evaluation perplexity of different data selection methods on WikiText-2 and WikiText-103.
Table~\ref{tab:results_reasoning} on the right compares accuracy on GSM8K and MATH.

\begin{table*}[t]
  \centering
  \small
  \setlength{\tabcolsep}{3pt}

  \begin{subtable}[t]{0.49\textwidth}
    \centering
    \caption{Language Modeling: Perplexity}
    \label{tab:results_clm}
    \resizebox{\linewidth}{!}{
    \begin{tabular}{lcccccc}
      \toprule
      \multirow{2}{*}{Selection Method} 
      & \multicolumn{3}{c}{WikiText-2} 
      & \multicolumn{3}{c}{WikiText-103} \\
      \cmidrule(lr){2-4} \cmidrule(lr){5-7}
      & [30\%] & [20\%] & [10\%] 
      & [30\%] & [20\%] & [10\%] \\
      \midrule

      \rowcolor{gray!15}
      \multicolumn{7}{c}{\textbf{LLaMA3.2-1B Base Model}} \\
      Full Fine-tuning & \multicolumn{3}{c}{8.79 [100\%]} 
                   & \multicolumn{3}{c}{8.00 [100\%]} \\
      Random & 9.21 & 9.37 & 9.69 & 8.18 & 8.28 & 8.41 \\
      Medium PPL & 9.22 & 9.36 & 10.52 & 8.22 & 8.30 & 8.45 \\
      High PPL & 9.25 & 9.37 & 9.63 & 8.30 & 8.38 & 8.51 \\
      \textbf{PPL-Factory} & \textbf{9.20} & \textbf{9.22} & \textbf{9.55} & \textbf{8.15} & \textbf{8.26} & \textbf{8.38} \\
      \midrule

      \rowcolor{gray!15}
      \multicolumn{7}{c}{\textbf{LLaMA3.2-3B Base Model}} \\
      Full Fine-tuning & \multicolumn{3}{c}{7.43 [100\%]} 
                   & \multicolumn{3}{c}{6.59 [100\%]} \\
      Random & 8.11 & 8.77 & 9.61 & 6.94 & 7.07 & 7.17 \\
      Medium PPL & 8.08 & 8.70 & 9.52 & \textbf{6.90} & 7.05 & 7.15 \\
      High PPL & 8.09 & 8.67 & 9.69 & 6.98 & 7.09 & 7.19 \\
      \textbf{PPL-Factory} & \textbf{8.04} & \textbf{8.64} & \textbf{9.43} & 6.92 & \textbf{7.02} & \textbf{7.13} \\
      \midrule

      \rowcolor{gray!15}
      \multicolumn{7}{c}{\textbf{LLaMA3.1-8B Base Model}} \\
      Full Fine-tuning & \multicolumn{3}{c}{6.14 [100\%]} 
                   & \multicolumn{3}{c}{5.90 [100\%]} \\
      Random & 6.32 & 6.40 & 6.66 & 6.14 & 6.20 & 6.37 \\
      Medium PPL & 6.32 & 6.42 & 6.64 & 6.16 & 6.19 & 6.34 \\
      High PPL & 6.32 & 6.40 & 6.61 & 6.29 & 6.33 & 6.40 \\
      \textbf{PPL-Factory} & \textbf{6.30} & \textbf{6.36} & \textbf{6.59} & \textbf{6.10} & \textbf{6.16} & \textbf{6.31} \\
      \bottomrule
    \end{tabular}
    }
  \end{subtable}
  \hfill
  \begin{subtable}[t]{0.49\textwidth}
    \centering
    \caption{Mathematical Reasoning: Accuracy}
    \label{tab:results_reasoning}
    \resizebox{\linewidth}{!}{
    \begin{tabular}{lcccccc}
      \toprule
      \multirow{2}{*}{Selection Method} 
      & \multicolumn{3}{c}{GSM8K} 
      & \multicolumn{3}{c}{MATH} \\
      \cmidrule(lr){2-4} \cmidrule(lr){5-7}
      & [30\%] & [20\%] & [10\%] 
      & [30\%] & [20\%] & [10\%] \\
      \midrule

      \rowcolor{gray!15}
      \multicolumn{7}{c}{\textbf{LLaMA3.2-1B-Instruct Model}} \\
      Full Fine-tuning & \multicolumn{3}{c}{39.88 [100\%]} 
                   & \multicolumn{3}{c}{19.4 [100\%]} \\
      Random & 36.62 & 36.16 & 35.94 & \textbf{21.0} & 19.8 & 19.8 \\
      Medium PPL & 35.41 & 34.19 & 34.95 & 19.0 & 17.2 & 20.4 \\
      High PPL & 35.71 & 34.34 & 31.92 & 20.2 & 21.2 & 21.0 \\
      \textbf{PPL-Factory} & \textbf{39.42} & \textbf{36.92} & \textbf{36.32} & 20.6 & \textbf{22.4} & \textbf{21.8} \\
      \midrule

      \rowcolor{gray!15}
      \multicolumn{7}{c}{\textbf{LLaMA3.2-3B-Instruct Model}} \\
      Full Fine-tuning & \multicolumn{3}{c}{61.87 [100\%]} 
                   & \multicolumn{3}{c}{27.2 [100\%]} \\
      Random & 62.40 & 59.89 & 62.55 & 26.0 & 27.6 & 28.8 \\
      Medium PPL & 62.77 & 61.20 & 59.89 & 26.2 & 25.4 & 24.2 \\
      High PPL & 59.89 & 60.50 & 59.89 & 27.4 & 29.4 & 29.0 \\
      \textbf{PPL-Factory} & \textbf{63.00} & \textbf{61.56} & \textbf{62.77} & \textbf{28.0} & \textbf{29.8} & \textbf{32.0} \\
      \midrule

      \rowcolor{gray!15}
      \multicolumn{7}{c}{\textbf{LLaMA3.1-8B-Instruct Model}} \\
      Full Fine-tuning & \multicolumn{3}{c}{71.42 [100\%]} 
                   & \multicolumn{3}{c}{36.2 [100\%]} \\
      Random & 69.06 & 66.79 & 64.67 & 35.6 & 34.4 & 35.2  \\
      Medium PPL & 65.50 & 65.20 & 65.13 & 31.6 & 33.4 & 33.2 \\
      High PPL & 63.76 & 59.44 & 58.76 & 33.8 & 32.0 & 34.6 \\
      \textbf{PPL-Factory} & \textbf{70.46} & \textbf{68.34} & \textbf{67.25} & \textbf{35.8} & \textbf{34.6} & \textbf{35.6} \\
      \bottomrule
    \end{tabular}
    }
  \end{subtable}
  \caption{Results of PPL-Factory under different data budgets. 
  Table(a) on the left reports perplexity on WikiText-2 and WikiText-103, where lower is better. 
  Table(b) on the right reports accuracy on GSM8K and MATH, where higher is better.
  Medium PPL and High PPL are reproduced from~\cite{ankner2025perplexed}, using deterministic categories.
  Best results within each column are in \textbf{bold}.
  Data selection ratios are in [brackets].}
  \label{tab:main_results}
\end{table*}

\paragraph{Causal Language Modeling Performance}
Overall, PPL-Factory provides the most stable performance across WikiText-2 and WikiText-103.
In nearly all settings, it achieves the lowest evaluation perplexity among the data selection baselines, demonstrating its effectiveness in identifying informative subsets for language-modeling fine-tuning.
Although full-data fine-tuning remains the upper reference in most cases, PPL-Factory substantially narrows the gap using only $10\%$ of the training data.
It shows the effectiveness of applying random selection to the middle band under low data budget settings.

\paragraph{Reasoning Tasks Accuracy}
Compared with existing perplexity-based methods, PPL-Factory achieves higher accuracy on both GSM8K and MATH under all the experimental settings.
In addition, when fine-tuning LLaMA3.2-3B-Instruct model on GSM8K with $30\%$ and $10\%$ of data, PPL-Factory yields better performance than the model fine-tuned on the entire training dataset.
Experimental results on MATH also demonstrate that PPL-Factory provides strong subset fine-tuning performance.
More specifically, for 1B and 3B models, fine-tuning on the selected subsets achieves higher accuracy than full fine-tuning across all the $30\%$, $20\%$, and $10\%$ selection ratios, indicating that task-aware selection can remove less useful samples while preserving more effective reasoning signals.

\begin{table}[t]
  \centering
  \small
  \setlength{\tabcolsep}{3pt}
  \resizebox{\columnwidth}{!}{%
  \begin{tabular}{lccc}
    \toprule
    \textbf{Method} & [$1\%$] & [$5\%$] & [$10\%$] \\
    \midrule
    Llama3.1-8B-Instruct & \multicolumn{3}{c}{57.49 [$0\%$]} \\
    \; + Full Fine-tuning & \multicolumn{3}{c}{71.39 [$100\%$]} \\
    \; + Random & 58.09 & 64.17 & 65.67 \\
    \; + Medium PPL~\cite{ankner2025perplexed} & 58.07 & 67.70 & 66.43 \\
    \; + High PPL~\cite{ankner2025perplexed} & 56.33 & 60.88 & 59.71 \\
    \; + GraNd~\cite{10.5555/3540261.3541836} & 57.23 & 67.65 & 70.05 \\
    \; + EL2N~\cite{10.5555/3540261.3541836} & 60.16 & 61.50 & 71.66 \\
    \; + CCS~\cite{zheng2023coveragecentric} & 60.70 & 63.64 & 68.72 \\
    \; + Data Whisperer~\cite{wang-etal-2025-data-whisperer} & 62.57 & \textbf{69.68} & \textbf{72.46} \\
    \; + \textbf{PPL-Factory (Ours)} & \textbf{62.94} & 69.61 & 70.24 \\
   \bottomrule
  \end{tabular}%
  }
  \caption{Exact-match accuracy on GSM8K dataset, using LLaMA3.1-8B-Instruct Model.
  [$0\%$] means the model is not fine-tuned on the downstream dataset.
  [$100\%$] refers to the model being fine-tuned with the full-size training set, without data selection.}
  \label{tab:results_compare}
\end{table}

Compared to other data selection methods, PPL-Factory achieves higher accuracy with only $1\%$ data, as shown in Table~\ref{tab:results_compare}.
Although using $10\%$ data does not achieve higher accuracy than EL2N and Data Whisperer, our PPL-Factory is computational resource-friendly and can be performed on a single NVIDIA H100 GPU.
EL2N is memory-intensive because its token-level prediction involves vocabulary-sized logits, whereas our NLL scoring can be computed with inference-only memory and scalar scores.
The attention-based Data Whisperer needs to be performed on eight NVIDIA A100 GPUs~\cite{wang-etal-2025-data-whisperer}.
Experimental results show that our task-aware NLL score works effectively on reasoning tasks, and especially fits the low data-budget settings.

\paragraph{Budget-Aware Selection Criterion}
The best data selection criterion shifts with the fine-tuning data budget.
Among the five categories, \texttt{random}, \texttt{easy}, \texttt{middle}, \texttt{mid\_random}, and \texttt{hard}, easy samples contribute the most learning signals when a relatively large portion of the training set is preserved.
These low-perplexity examples can provide stable and reliable weight updates, making selection effective.
In contrast, when only a very limited subset is kept, i.e., less than $30\%$ of the full-size dataset, deterministic selection from a narrow, biased perplexity region may reduce data diversity and over-concentrate on similar training examples.
Under this condition, \texttt{mid\_random}, which focuses on a wider middle band and selects randomly from the candidate pool, achieves a higher expected utility score and thus is the preferred selection criterion with a better evaluation performance.
In other words, introducing randomness enhances the diversity of the selected data, and contributes positively to fine-tuning.
It balances training effectiveness and efficiency without involving extra computational cost overhead.
Thus, we conclude that the data selection method with the best performance is not only determined by the model size or the tasks, but also can be versatile, given the percentage of the training samples.

\subsection{Ablation}
\label{sec:exp_results_sub_ablation}
Table~\ref{tab:results_reasoning_ablation} presents the GSM8K accuracy of our combined NLL score in comparison to different NLL scoring variants.
The results show that PPL-Factory consistently achieves the highest exact-match accuracy across all selection ratios.
Compared with random selection, PPL-Factory improves the accuracy by $2.80$, $0.76$, and $0.38$ points using $30\%$, $20\%$, and $10\%$ of training samples, respectively.
Among the single-component variants, reasoning NLL and answer NLL can provide competitive performance with $30\%$ data, but their performance becomes less stable when the data budget decreases.
Response NLL also shows limited improvement, suggesting that using the full response as a single scoring target may dilute the contribution of the reasoning and final-answer components.
In contrast, PPL-Factory combines both reasoning steps and answer information and achieves the best performance in all three settings, indicating that both components in Equation~\ref{eq:nll_reasoning} are useful for selecting informative reasoning examples.

\begin{table}[t]
  \centering
  \small
  \setlength{\tabcolsep}{3pt}
  \resizebox{\columnwidth}{!}{%
  \begin{tabular}{lccc}
    \toprule
    \textbf{Method} & [$30\%$] & [$20\%$] & [$10\%$] \\
    \midrule
    Random & 36.62 & 36.16 & 35.94 \\
    Response NLL & 36.47 & 36.39 & 35.25 \\
    Reasoning NLL & 38.97 & 35.48 & 36.01 \\
    Answer NLL & 38.67 & 34.34 & 34.72 \\
    \textbf{PPL-Factory (Ours)} & \textbf{39.42} & \textbf{36.92} & \textbf{36.32} \\
   \bottomrule
  \end{tabular}%
  }
  \caption{Ablation of our PPL-Factory compared with NLL variants fine-tuned on GSM8K using LLaMA3.2-1B-Instruct Model.
  Response NLL use the full response to calculate the selection score.
  Reasoning NLL is equivalent to set controlled weight in Equation~\ref{eq:nll_reasoning} to $\alpha=1$,\, $\beta=0$, and answer NLL is equivalent to set $\alpha=0$,\, $\beta=1$.}
  \label{tab:results_reasoning_ablation}
\end{table}

\subsection{Calculation Complexity and Training Cost}
\label{sec:exp_results_sub_complexity}
In the PPL-Factory framework, for a dataset with $N$ examples, each NLL-based score is computed by forward inference using the target model, where the subset is directly determined by the target model without using proxy evaluation.
The main efficiency gain comes from reducing the size of the training set. 
With a selection ratio $\rho \in (0, 1]$, the number of training examples is reduced from $N$ to $\rho N$, and the fine-tuning cost decreases approximately linearly with the number of processed tokens.
Although the scoring step adds a one-time inference overhead, this cost is much smaller than the cost of full fine-tuning. 
Experimental results show that our method achieves a trade-off between performance and training cost.

\section{Related Work}
Data selection aims to identify a smaller but more useful subset that preserves downstream performance.
Existing methods can be broadly categorized into difficulty or perplexity-based selection, coverage-based selection, and influence-based selection.

\paragraph{Difficulty- or Perplexity-based Selection}
Some data pruning methods, such as GraNd and EL2N~\cite{10.5555/3540261.3541836}, estimate the importance of examples from early training dynamics by the L2 norm of error vectors.
RHO-loss~\cite{pmlr-v162-mindermann22a} prioritizes examples that are learnable, useful, but not yet learned.
$\mathcal{H}$-score~\cite{pmlr-v274-hosseini25a} ranks each training example by the success rate in classifying it correctly across multiple fine-tuning runs.

Perplexity-based data pruning has been shown to be effective for pre-training~\cite{marion2023moreinvestigatingdatapruning}.
Recent work~\cite{ankner2025perplexed} demonstrates that a small reference model can provide a useful perplexity-based rank for selecting examples, which can be used to train larger models.
These methods suggest that model-estimated difficulty can serve as a proxy estimation for data utility.

\paragraph{Coverage-based Selection}
Coverage-centric Coreset Selection~\cite{zheng2023coveragecentric} argues that maintaining coverage is important under high pruning rates.
The existing selection algorithms may over-concentrate a narrow region of the data distribution and suffer from a dramatic accuracy drop under extremely low data-budget settings.

\paragraph{Influence-based Selection}
LESS~\cite{10.5555/3692070.3694291} selects influential examples for target instruction tuning by comparing low-dimensional gradient features between candidate examples and target-task demonstrations.
Influence-based pruning~\cite{humane2025influence, li-etal-2024-quantity} improves reasoning-data selection over heuristic criteria, but it requires costly influence estimation and shows limited capability in cross-model transferability.

\paragraph{Our PPL-Factory Framework}
Different from methods that rely on external LLM references and expensive gradient similarity, our work studies a simple and scalable one-shot data selection framework based on task-aware NLL scores and budget-aware selection criteria.
Our study connects perplexity-based data selection approaches with supervised fine-tuning for both CLM tasks and mathematical reasoning tasks.
Empirical evidence shows that our PPL-Factory outperforms deterministic perplexity-based selection and other high-cost approaches, including gradient-based GraNd and EL2N, coverage-based CCS, and attention-based Data Whisperer, while keeping the simplicity and low-resource requirements.

\section{Conclusion}
\label{sec:conclusion}
In this paper, we presented PPL-Factory, a task-aware and budget-aware data selection framework for efficient fine-tuning of large language models.
Experiments on CLM benchmarks show that PPL-Factory preserves most of the full-data fine-tuning performance using only $10\%$ of the training corpus.
On GSM8K, PPL-Factory achieves higher accuracy than state-of-the-art selection methods using $1\%$ of the training data.
It also surpass full-data fine-tuning in several low-budget settings on both reasoning benchmarks, GSM8K and MATH, by removing examples with weak or noisy learning signals.
Overall, PPL-Factory reduces training cost across both language modeling and reasoning tasks by integrating the simplicity of the perplexity-based scores adapted to different task structures with budget-aware selection strategies.

\section*{Limitations}
While our work demonstrates the promise of offline one-shot data selection for both causal language modeling and mathematical reasoning, several limitations remain.
Our study mainly focuses on supervised fine-tuning, where reducing the number of training examples directly lowers computational cost, especially when the dataset size dominates the overall training expense.
We leave the extension of PPL-Factory to knowledge distillation as future work.
In particular, applying perplexity-based selection to teacher–student learning may further reduce training cost by enabling smaller student models to learn from more informative subsets of the distillation data.
With an appropriate student architecture, we believe PPL-Factory can provide an efficient solution for improving knowledge transfer while using fewer training samples.

\bibliography{custom}

\appendix

\section{Overview}
\label{sec:appendix_overview}

This appendix provides details of our theoretical analysis, experimental settings and additional evaluation results.

\section{Theoretical Analysis}
\label{sec:appendix_maximizer}
\subsection{Proof of the Optimizer in Equation~\ref{eq:informative_coverage_maximizer}}
We consider the informative-coverage score defined in Section~\ref{sec:method_sub_problem}:
\begin{equation}
  \begin{aligned}
    J_\rho(Q_\rho) &= \int u_\rho(s) Q_\rho(s) \, ds \\
    &- \lambda_\rho D_{\mathrm{KL}}(Q_\rho \parallel P), & \lambda_\rho > 0
  \end{aligned}
  \label{eq:score_informative_coverage_appendix}
\end{equation}
where $u_\rho(s)$ denotes the expected utility of retaining an example with score $s$ under the selecting ratio $\rho$, and $\lambda_\rho > 0$ controls the preference for preserving coverage relative to concentrating on high-utility regions.

The function $u_\rho(s)$ is deliberately budget-dependent.
Intuitively, when $\rho \in (0, 1]$ is relatively large, the selected subset already preserves most of the original data distribution, so the main benefit comes from removing a small, low-quality tail.
When $\rho$ is relatively small, however, each retained example must carry more learning value, and the optimal selector should focus more strongly on informative but still learnable examples.

\paragraph{Proposition}
\label{prop:budget_optimal_selector}
For a fixed ratio $\rho$, the unique maximizer of Equation~\ref{eq:informative_coverage_maximizer} is
\begin{equation}
    Q_\rho^*(s) = \frac{P(s) \exp(u_\rho(s)/\lambda_\rho)}{\int P(t) \exp(u_\rho(t)/\lambda_\rho) \, dt}
  \label{eq:informative_coverage_maximizer_appendix}
\end{equation}

\begin{proof}
Consider the Lagrangian
\begin{equation}
  \begin{aligned}
    \mathcal{L}(Q_\rho,\eta) &= \int u_\rho(s)Q_\rho(s)\, ds \\
    &- \lambda_\rho \int Q_\rho(s)\log\frac{Q_\rho(s)}{P(s)}\, ds \\
    &+ \eta\left(\int Q_\rho(s)\, ds - 1\right)
  \end{aligned}
  \label{eq:appendix_lagrangian}
\end{equation}

Take the derivative with respect to $Q_\rho(s)$, and make it zero gives
\begin{equation}
    u_\rho(s) - \lambda_\rho \left(\log\frac{Q_\rho(s)}{P(s)} + 1\right) + \eta = 0
  \label{eq:appendix_derivative}
\end{equation}

It yields
\begin{equation}
    Q_\rho(s) = P(s)\exp(u_\rho(s)/\lambda_\rho) \, C,
\end{equation}
for normalization constant $C$.

Enforcing $\int Q_\rho(s)\, ds = 1$ yields the maximizer~\ref{eq:informative_coverage_maximizer_appendix}. Since the objective in Equation~\ref{eq:score_informative_coverage_appendix} is strictly concave in $Q_\rho$, the solution is unique.
\end{proof}
Thus, the data selection is to find a subset of the training dataset which is closer to the optimizer $Q_\rho(s)$.

\subsection{Discussion on Pruning Rates}
\begin{figure*}[t]
  \includegraphics[width=0.48\linewidth]{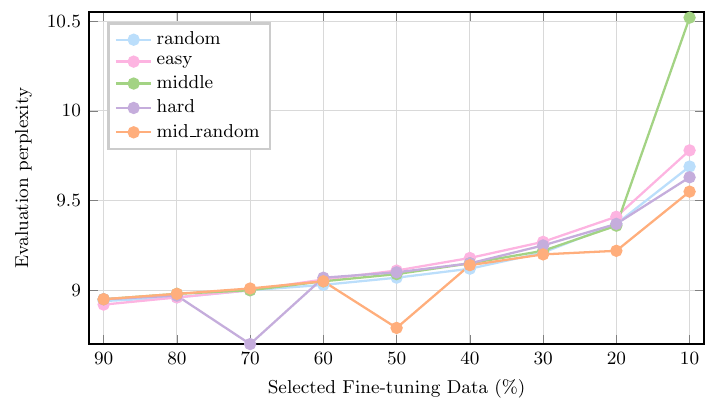} \hfill
  \includegraphics[width=0.48\linewidth]{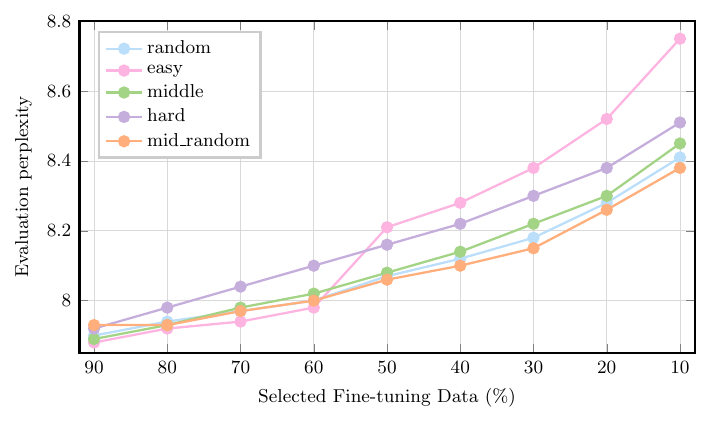}
  \centering
  \caption{Comparison of the evaluation perplexity using different subsets from WikiText-2 (left) and WikiText-103 (right) dataset, using LLaMA3.2-1B model.}
  \label{fig:nll_selection_1B}
\end{figure*}

\begin{figure*}[t]
  \includegraphics[width=0.48\linewidth]{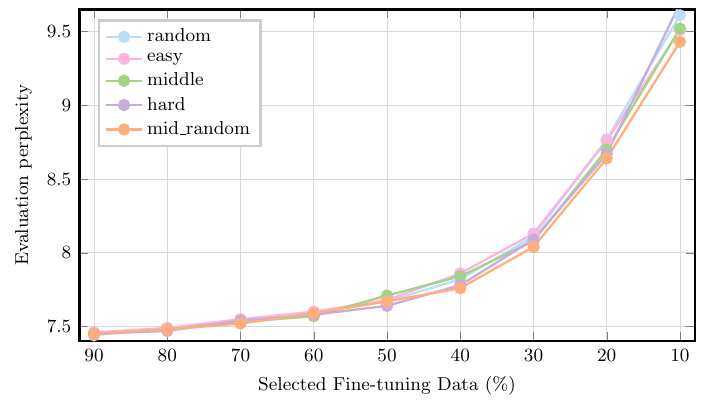} \hfill
  \includegraphics[width=0.48\linewidth]{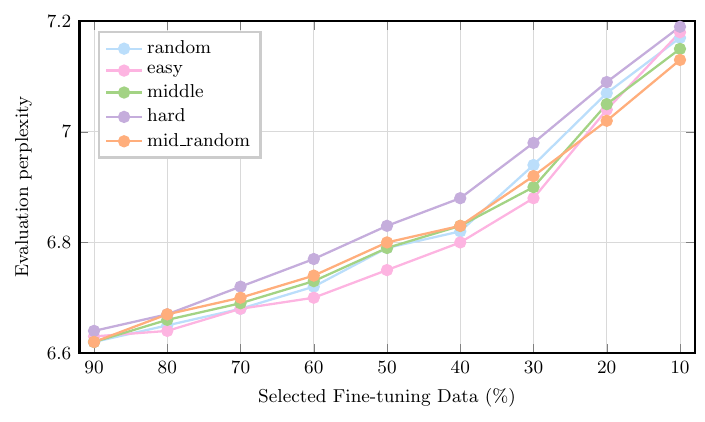}
  \centering
  \caption{Comparison of the evaluation perplexity using different subsets from WikiText-2 (left) and WikiText-103 (right) dataset, using LLaMA3.2-3B model.}
  \label{fig:nll_selection_3B}
\end{figure*}

\begin{figure*}[t]
  \includegraphics[width=0.48\linewidth]{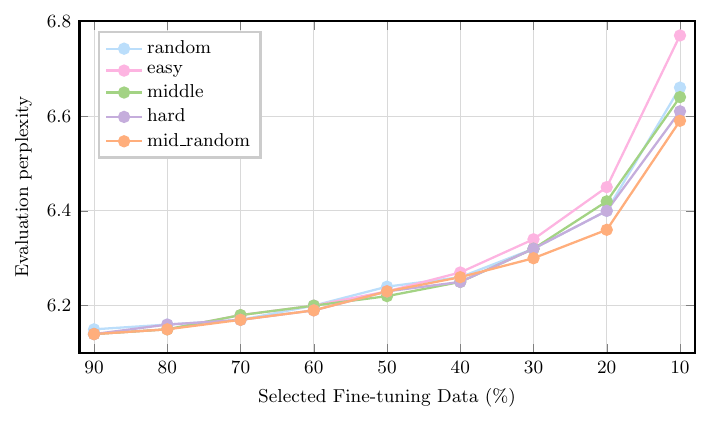} \hfill
  \includegraphics[width=0.48\linewidth]{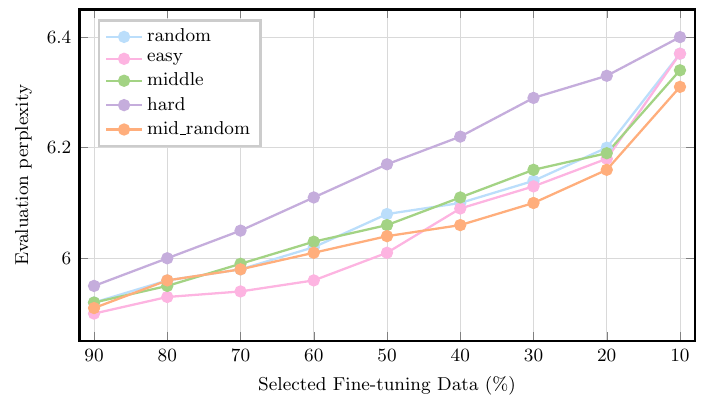}
  \centering
  \caption{Comparison of the evaluation perplexity using different subsets from WikiText-2 (left) and WikiText-103 (right) dataset, using LLaMA3.1-8B model.}
  \label{fig:nll_selection_8B}
\end{figure*}

The shape of the optimal selector is controlled by the budget-dependent utility $u_\rho(s)$. 
Thus, the preferred NLL region may shift with the size of available training samples.
We compare the evaluation perplexity on WikiText-2 and WikiText-103 using different selection ratios on training samples in Figure~\ref{fig:nll_selection_1B} to Figure~\ref{fig:nll_selection_8B}.

\paragraph{High Selection Ratio}
When using a relatively higher selection ratio, for example, retaining $60\% $--$90\%$ of the training data, selecting \texttt{easy} examples with a lower NLL score can outperform the other four categories, \texttt{random}, \texttt{middle}, \texttt{mid\_random}, and \texttt{hard}.
However, the kept subset remains close in distribution to the full dataset. 
The main advantage of selection under this circumstance is often the removal of the hardest tail in a low-density region, which may contain noisy, atypical, or low-return examples.

\paragraph{Low Selection Ratio}
Under a very limited computational budget, i.e., to force the selection ratio to be lower, middle-band selection with randomization can become more effective.
Under this condition, training samples with a low NLL score contains limited learning signal that would contribute to the training and weight updates.
Thus, these examples may become redundant, while examples with very high NLL scores may remain too noisy or difficult.
The best trade-off can then shift to the middle region, but it still needs the diversity from randomness.

\paragraph{Conclusion}
The selection method with the best performance is not only determined by the model size or the tasks, but also can be versatile, given the percentage of the kept training samples.
The framework above suggests that there is no universally optimal fixed NLL region for data selection.
Instead, the preferred region is budget-dependent:
\begin{equation}
    Q_\rho^\star(s) \propto P(s) \exp(u_\rho(s)/\lambda_\rho).
\end{equation}
For a high selection ratio, $u_\rho(s)$ may favor easy examples, making low-NLL retention optimal.
For a low selection ratio, $u_\rho(s)$ may peak in the middle, making middle-biased randomized selection preferable.

\section{Reproducibility Details}
\label{sec:appendix_repro}
\subsection{Data Split}
In this section, we are going to introduce the benchmarks used in our experiments in detail.
The train set and validation set split is summarized in Table~\ref{tab:dataset_splits}.
\begin{table*}[t]
  \centering
  \small
  \resizebox{\linewidth}{!}{
  \begin{tabular}{cccc}
    \toprule
    \textbf{Dataset} & \textbf{Selection Pool} & \textbf{Evaluation Set} & \textbf{Metric} \\
    \midrule
    \makecell[c]{WikiText-2} & \makecell[c]{WikiText-2 train \\ (36,718 samples)}  & \makecell[c]{WikiText-2 validation \\ (3,760 samples)} & Perplexity \\
    \makecell[c]{WikiText-103}
    & \makecell[c]{WikiText-103 train \\ (1,801,350 samples)}
    & \makecell[c]{WikiText-103 validation \\ (3,760 samples)} & Perplexity \\
    \makecell[c]{GSM8K}
    & \makecell[c]{GSM8K train \\ (7,473 samples)}
    & \makecell[c]{GSM8K test \\ (1,319 samples)} & Exact Match \\
    \makecell[c]{MATH}
    & \makecell[c]{MATH-minus-MATH500 \\ (12,000 samples)}
    & \makecell[c]{MATH500 \\ (500 samples)} & Greedy Pass@1 \\
    \bottomrule
  \end{tabular}
  }
  \caption{Dataset splits used in our experiments.}
  \label{tab:dataset_splits}
\end{table*}

\paragraph{WikiText-2 and WikiText-103}
For language modeling experiments, we use the official WikiText-2 and WikiText-103 splits.
The WikiText language modeling dataset is a collection of tokens extracted from Wikipedia.
The training split is used for data selection and subsequent fine-tuning.
Specifically, the text is tokenized, concatenated, and divided into fixed-length token blocks, and selection is performed at the block level. 
The official validation and test splits are kept separate from the selection process and are used only for perplexity evaluation. 
WikiText-2 contains 36,718 training lines, 3,760 validation lines, and 4,358 test lines, while WikiText-103 contains 1,801,350 training lines, 3,760 validation lines, and 4,358 test lines.

\paragraph{GSM8K}
GSM8K (Grade School Math 8K) is a dataset of 8.5K high-quality, linguistically diverse grade school math word problems.
For GSM8K, we use the official training split for data selection and fine-tuning. 
The official test split is used only for evaluation. 
The GSM8K training split contains 7,473 examples, and the test split contains 1,319 examples. 
We report zero-shot exact-match accuracy on the evaluation set.

\paragraph{MATH}
The MATH benchmark~\cite{hendrycksmath2021} has 12,500 competition-level math problems with step-to-step solutions.
MATH500~\cite{lightman2024lets} comes from the nonstandard MATH split, where 500 held-out problems are used for evaluation.
For MATH data selection and fine-tuning, we use MATH-minus-MATH500, a training pool containing 12,000 examples derived from MATH by removing the 500 MATH500 problems.
Following common practice, we evaluate the fine-tuned models on MATH500.
No MATH500 examples are included in the selected training subset.

\subsection{Fine-tuning Details}
\label{sec:appendix_repro_sub_finetuning}
The learning rate during fine-tuning is set to $2e-5$ for all the 1B/3B/8B models.
In the experiments on WikiText-2 and WikiText-103, we apply evaluation during training and save the best model with the lowest evaluation perplexity.
For data budget {$10\%$, $20\%$} on WikiText-2, we run 2 epochs to avoid overfitting.
For the rest of the data budget, we run 3 epochs.
In the experiments on GSM8K and MATH, we perform evaluation after fine-tuning.
All experiments are conducted for 3 epochs.

\end{document}